%% file: root.tex
\documentclass[letterpaper, 10 pt, conference]{ieeeconf}

\IEEEoverridecommandlockouts                              

\usepackage{amsmath, scalerel} 
\usepackage{amssymb}
\usepackage{amsopn}
\usepackage{graphicx}
\usepackage[style=base]{caption}
\usepackage{color}
\usepackage{quotmark}
\usepackage{booktabs}
\usepackage{todonotes}
\usepackage{lipsum}
\usepackage{import}
\usepackage{footnote}
\usepackage{multirow}
\usepackage{hhline}
\usepackage{subcaption}
\usepackage[pdfborder={0 0 0}, bookmarks=false, final]{hyperref}

\usepackage{floatrow}

\usepackage[abbreviations,binary-units]{siunitx}
\DeclareSIUnit{\mAh}{mAh}
\DeclareSIUnit{\Wh}{Wh}

\usepackage{tikz}
\usetikzlibrary{calc}
\usetikzlibrary{matrix}
\usetikzlibrary{math}
\usetikzlibrary{plotmarks}
\usetikzlibrary{positioning}
\usetikzlibrary{shapes}
\usetikzlibrary{arrows}
\usetikzlibrary{arrows.meta}

\usepackage{pgfplots}
\pgfplotsset{compat=newest}
\usepgfplotslibrary{patchplots}
\usepackage{grffile}

\usepackage{pgfplots}
\pgfplotsset{compat=newest} 
\pgfplotsset{plot coordinates/math parser=false} 
\newlength\figureheight 
\newlength\figurewidth 

\title{\LARGE \bf
Path Evaluation via HMM on Semantical Occupancy Grid Maps
}

\author{Timo Korthals$^{*}$, Julian Exner, Thomas Sch{\"o}pping, and Marc Hesse
\thanks{Bielefeld University, Cluster of Excellence Cognitive Interaction Technologies, Cognitronics \& Sensor Systems,
        Inspiration 1, 33619 Bielefeld, Germany,
        {\tt\small http://www.ks.cit-ec.uni-bielefeld.de/},
        {$^{*}$\tt\small tkorthals@cit-ec.uni-bielefeld.de}}
}

\begin{document}

\maketitle
\thispagestyle{empty}
\pagestyle{empty}

\begin{abstract}
Traditional approaches to mapping of environments in robotics make use of spatially discretized representations, such as occupancy grid maps.
Modern systems, e.g.\ in agriculture or automotive applications, are equipped with a variety of different sensors to gather diverse process-relevant modalities from the environment.
The amount of data and its associated semantic information demand for broader data structures and frameworks, like semantical occupancy grid maps (SOGMs).
This multi-modal representation also calls for novel methods of path planning.
Due to the sequential nature of path planning as a consecutive execution of tasks and their ability to handle multi-modal data as provided by SOGMs, Markovian models, such as Hidden Markov Models (HMM) or Partially Observable Markov Decision Processes, are applicable.
Furthermore, for these techniques to be applied effectively and efficiently, data from SOGMs must be extracted and refined.
Superpixel algorithms, originating from computer vision, provide a method to de-noise and re-express SOGMs in an alternative representation.
This publication explores and extends the use of superpixel segmentation as a post-processing step and applies Markovian models for path decoding on SOGMs.
\end{abstract}
\section{INTRODUCTION}
Decoding refers to inferring the true state of the world that led to a specific observation, i.e. the value of a semantical occupancy grid cell.
Depending on the application there are different states of interest.
A basic scenario in mobile robotics is obstacle avoidance, where occupancyness is the objective state.
But more sophisticated scenarious in which mobile robots process their environment, like harvesting machines or robotic vacuum cleaners, call for a more diverse set of states.
On one hand, this motivates the introduction of SOGMs but also raises the question how SOGM cells may be mapped to states of the real world.
This may be described as a classification problem, in which cells are assigned to classes, i.e. an underlying state of the world, based on the associated probabilities in the SOGM.
However, since real-world environments are usually highly structured, cells in an SOGM should not be seen as independent w.r.t their class or spatial relation.
In mobile, and especially embedded, robotics only the states along a specific trajectory are of interest due to keep the computational costs low.
The combination of these two considerations suggests the use of Hidden Markov Models (HMM) to decode the sequence of hidden states along a proposed trajectory.
When applying HMMs to this problem, it is formulated such that each hidden state is mapped to a class.
\section{APPROACH}
\subsection{ENVIRONMENT \& PERCEPTION MODELS}
\begin{figure}
    \tiny
    \raisebox{-\height}{\includegraphics[width=0.42\linewidth]{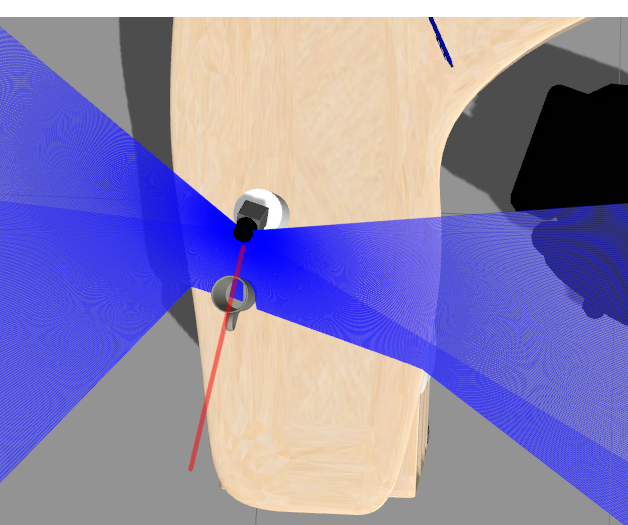}}
	\hspace{2em}
	\raisebox{-\height}{\input{graph_object.tex}}
	\caption[Behaviour of SOGMs when approaching an object]{Top view of a table-top robot approaching a mug in simulation (left).
		Complex behavior of the individual SOGM layers along the red line (right).
		}
	\label{fig:hmm-graph-object}
\end{figure}

Famous representation of the environment are 2D or 3D \textit{occupancy grids} $M$ introduced by Moravec and Elfes \cite{Moravec1985}, where each cell $m\in M$ is associated with a Bernoulli random variable indicating the probability $P(m)$ of occupation.
For any observed cell $m$ at time $t$, a log-likelihood update of the odds probability can be performed, given the observers position $x$ and observation $z$:
$L(m | z_{1:t}, x_{1:t}) = L(m | z_{1:{t-1}}, x_{1:t-1}) + L(m |z_t, x_t)$,
where $L(m | z_t, x_t)$ denotes the \textit{inverse sensor model} capturing a classifiers output, sensor calibration, and registration.
An overview of various model designs for the agricultural domain is given in \cite{Korthals2018}.
While a single occupancy grid is only able to represent one modality, a functional extension comprising $N$ modalities, with $\mathbf{P}(m)=\left(P_1(m), \ldots, P_N(m) \right)$ as \textit{semantical occupancy grids}, is introduced in \cite{Korthals2017a}.
For various path evaluation or traversal applications working on a single modal occupancy grid, cell-wise consideration
might be feasible.
However, multi-modal semantical grids do suffer from noisy or sparse data and potential registration errors between sensors.
Thus, map refinement via clustering was introduced by \cite{Korthals2017a} using a \textit{Supercell Extracted Variance Driven Sampling} algorithm, whose objective is to find clusters that consist of non-contradicting cell probabilities (cf. \autoref{fig:hmm-graph-object}).
\subsection{PATH DECODING}
\begin{figure}
    \footnotesize
\begingroup
  \makeatletter
  \providecommand\color[2][]{
    \errmessage{(Inkscape) Color is used for the text in Inkscape, but the package 'color.sty' is not loaded}
    \renewcommand\color[2][]{}
  }
  \providecommand\transparent[1]{
    \errmessage{(Inkscape) Transparency is used (non-zero) for the text in Inkscape, but the package 'transparent.sty' is not loaded}
    \renewcommand\transparent[1]{}
  }
  \providecommand\rotatebox[2]{#2}
  \newcommand*\fsize{\dimexpr\f@size pt\relax}
  \newcommand*\lineheight[1]{\fontsize{\fsize}{#1\fsize}\selectfont}
  \ifx\svgwidth\undefined
    \setlength{\unitlength}{271.63751221bp}
    \ifx\svgscale\undefined
      \relax
    \else
      \setlength{\unitlength}{\unitlength * \real{\svgscale}}
    \fi
  \else
    \setlength{\unitlength}{\svgwidth}
  \fi
  \global\let\svgwidth\undefined
  \global\let\svgscale\undefined
  \makeatother
  \begin{picture}(1,0.31268438)
    \lineheight{1}
    \setlength\tabcolsep{0pt}
    \put(0,0){\includegraphics[width=\unitlength,page=1]{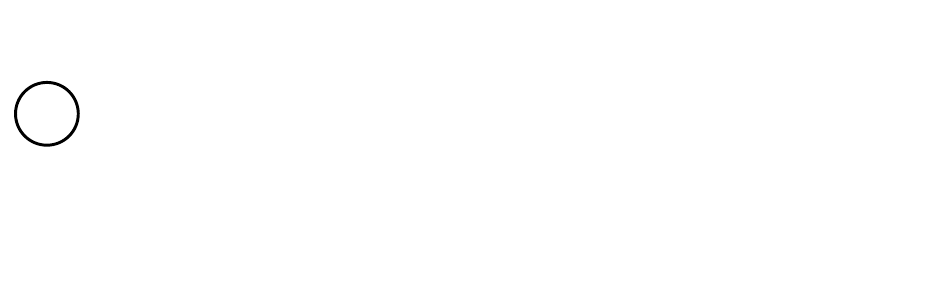}}
    \put(0.03118586,0.18571244){\color[rgb]{0,0,0}\makebox(0,0)[lt]{\lineheight{0}\smash{\begin{tabular}[t]{l}$w_2$\end{tabular}}}}
    \put(0,0){\includegraphics[width=\unitlength,page=2]{HMM_Process.pdf}}
    \put(0.12065298,0.14670983){\color[rgb]{0,0,0}\makebox(0,0)[lt]{\lineheight{0}\smash{\begin{tabular}[t]{l}$\Phi_{w_2,J}$\end{tabular}}}}
    \put(0,0){\includegraphics[width=\unitlength,page=3]{HMM_Process.pdf}}
    \put(0.20118989,0.29478367){\color[rgb]{0,0,0}\makebox(0,0)[lt]{\lineheight{0}\smash{\begin{tabular}[t]{l}Trajectory\end{tabular}}}}
    \put(0,0){\includegraphics[width=\unitlength,page=4]{HMM_Process.pdf}}
    \put(0.2855389,0.14765798){\color[rgb]{0.49803922,0.49803922,0.49803922}\makebox(0,0)[lt]{\lineheight{0}\smash{\begin{tabular}[t]{l}\textit{Hidden Properties}\end{tabular}}}}
    \put(0.26027909,0.10969058){\color[rgb]{0.49803922,0.49803922,0.49803922}\makebox(0,0)[lt]{\lineheight{0}\smash{\begin{tabular}[t]{l}\textit{Observed SOGM Cluster}\end{tabular}}}}
    \put(0,0){\includegraphics[width=\unitlength,page=5]{HMM_Process.pdf}}
    \put(0.19142416,0.18645107){\color[rgb]{0,0,0}\makebox(0,0)[lt]{\lineheight{0}\smash{\begin{tabular}[t]{l}$w_3$\end{tabular}}}}
    \put(0.59403372,0.1855041){\color[rgb]{0,0,0}\makebox(0,0)[lt]{\lineheight{0}\smash{\begin{tabular}[t]{l}$w_1$\end{tabular}}}}
    \put(0,0){\includegraphics[width=\unitlength,page=6]{HMM_Process.pdf}}
    \put(0.03611967,0.26430778){\color[rgb]{0,0,0}\makebox(0,0)[lt]{\lineheight{0}\smash{\begin{tabular}[t]{l}$U$\end{tabular}}}}
    \put(0,0){\includegraphics[width=\unitlength,page=7]{HMM_Process.pdf}}
    \put(0.75457615,0.26772132){\color[rgb]{0,0,0}\makebox(0,0)[lt]{\lineheight{0}\smash{\begin{tabular}[t]{l}$U$\end{tabular}}}}
    \put(0.75456413,0.04035643){\color[rgb]{0,0,0}\makebox(0,0)[lt]{\lineheight{0}\smash{\begin{tabular}[t]{l}$\mathbf{P}$\end{tabular}}}}
    \put(0.79159416,0.00911781){\color[rgb]{0,0,0}\makebox(0,0)[lt]{\lineheight{0}\smash{\begin{tabular}[t]{l}$I$\end{tabular}}}}
    \put(0.87017393,0.14771392){\color[rgb]{0,0,0}\makebox(0,0)[lt]{\lineheight{0}\smash{\begin{tabular}[t]{l}$\lambda$\end{tabular}}}}
    \put(0.75211665,0.14900984){\color[rgb]{0,0,0}\makebox(0,0)[lt]{\lineheight{0}\smash{\begin{tabular}[t]{l}$w$\end{tabular}}}}
    \put(0,0){\includegraphics[width=\unitlength,page=8]{HMM_Process.pdf}}
    \put(0.88132378,0.09710735){\color[rgb]{0,0,0}\makebox(0,0)[lt]{\lineheight{0}\smash{\begin{tabular}[t]{l}$J$\end{tabular}}}}
    \put(0,0){\includegraphics[width=\unitlength,page=9]{HMM_Process.pdf}}
  \end{picture}
\endgroup

	\caption{Generative sampling (left) and model (right).
		}
	\label{fig:model}
\end{figure}

For the given task of path traversal, a hierarchical approach is targeted that not only models the single class at a certain location, but also the whole object itself.
The probability of observing the sequence $\mathcal{O}=\left( \mathbf{P}_1, \ldots, \mathbf{P}_{IJ}\right)$ with property $w$ can be expressed as the joint probability (cf. \autoref{fig:model}):
\begin{equation}
\label{eq:joint_prob}
P\left(\mathcal{O},w ; \lambda \right) = \prod^{I}_{i=1} P\left(w_{i}\right) \prod_{j=1}^{J} P\left(\mathbf{P}_{j} | w_i ; \lambda \right)\textrm{,}
\end{equation}
with the hidden variable $w$, $P(w)$ being the discrete property probability, and $\lambda$ being the generative property model for the observed feature vector $\mathcal{O}$.
The amount of properties along a path are enumerated by $I$ while the length of a single property is denoted by $J$.
The inter-property model $\lambda_{w}=\left(S, \mathcal{O}, A, \Phi, \Pi\right)$ is a corresponding Hidden Markov Model (HMM) with states $s$, observations $\mathcal{O}$, transition probability $A$, emission probability $\Phi$, and start probability $\Pi$ for every single property $w$.
The emission probability is modeled as a Gaussian Mixture Model (GMM) over the $N$ semantical occupancy grids with the assumption that the probabilities are Logit-normal distributed.

\section{EXPERIMENTS}
\subsection{SETUP}
\begin{figure}
\floatbox[{\capbeside\thisfloatsetup{capbesideposition={right,top},capbesidewidth=0.4\linewidth}}]{figure}[\FBwidth]
{\caption{Top: the AMiRo. Bottom: the simulated model of the AMiRo with tilted laser rangefinder and a front-facing camera indicated by the black dot.}\label{fig:amiro}}
{\includegraphics[width=1.05\linewidth]{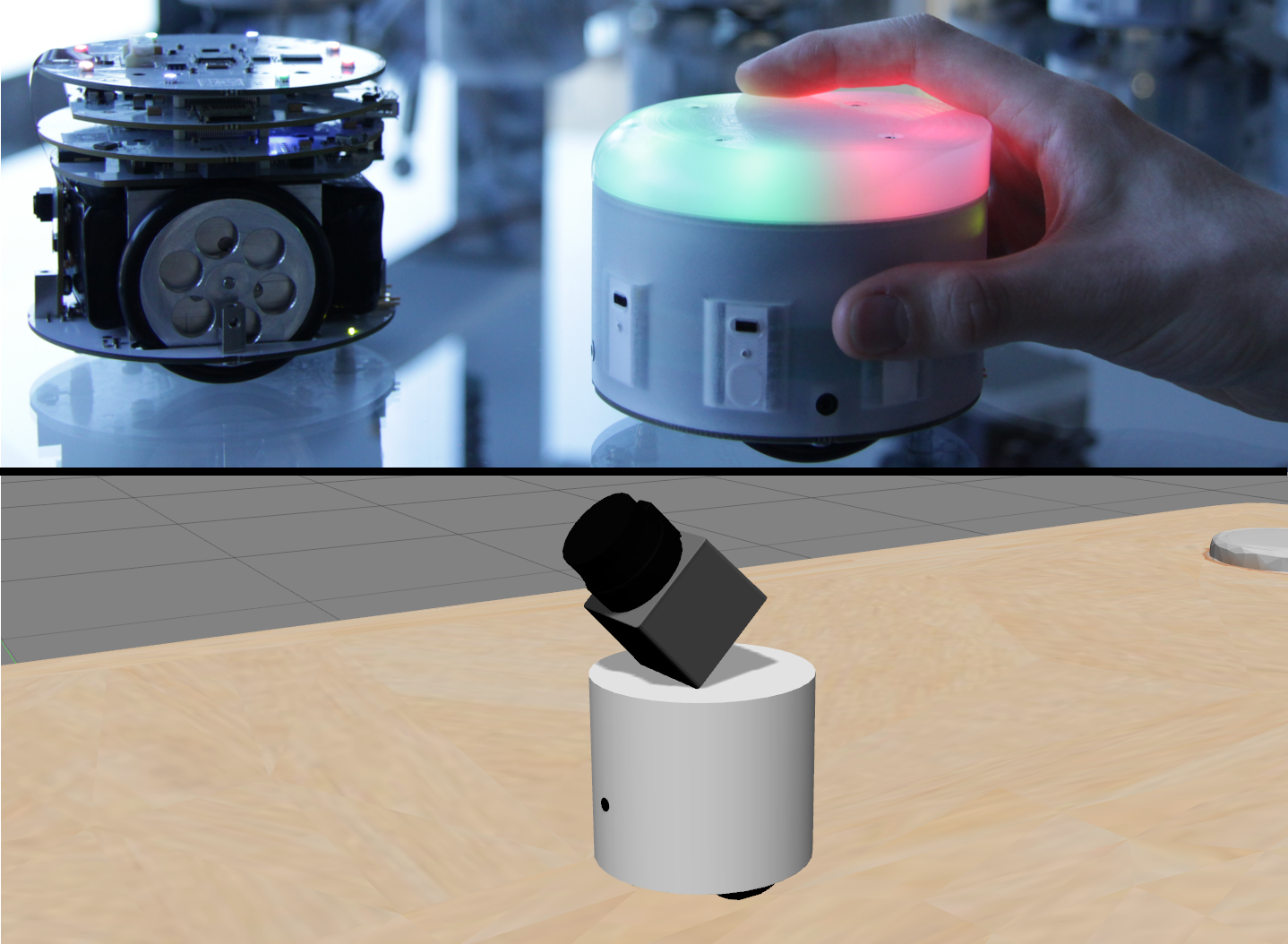}}
\end{figure}

The experiments were performed using the Autonomous Mini-Robot (AMiRo) in the corresponding Gazebo based simulation\footnote{\url{https://github.com/tik0/amiro_robot}} (cf. \autoref{fig:amiro}) with the goal of detecting classes along a planned trajectory while building the semantical occupancy grid in a table-top scenario.
Three classifiers\footnote{camera based anomaly detection via a neuronal network, LiDAR based obstacle and corner detection via a fuzzy classifier as introduced in \cite{Korthals2017a}}, and thus $N=3$ semantical layers, where applied.
Training and ground-truth data were recorded in Gazebo.
The HMM for path decoding was designed as left-right Bakis structure.
It was trained via \textit{hmmlearn}\footnote{\url{https://github.com/hmmlearn/hmmlearn}} in an unsupervised fashion to learn the classes \textit{ground}, \textit{table}, and \textit{object} on various pre-mapped table-top scenarios.
\subsection{RESULTS}

\begin{figure}
    \tiny
	\begin{subfigure}[t]{0.45\linewidth}
\begin{tikzpicture}
\footnotesize
\begin{axis}[
width=0.8\linewidth,
height=0.6\linewidth,
scale only axis,
unbounded coords=jump,
xmin=0.5,
xmax=3.5,
xtick={1,2,3},
xticklabels={{clustered},{cell-wise},{point cloud}},
xticklabel style={rotate=15, anchor=north east},
ymin=0.431086344259091,
ymax=0.685133966540419,
ylabel style={font=\color{white!15!black}},
ylabel={$\text{F}_\text{1}$},
axis background/.style={fill=white},
xmajorgrids,
ymajorgrids
]
\addplot [color=black, dashed, forget plot]
  table[row sep=crcr]{
1	0.658153960883433\\
1	0.673586347345813\\
};
\addplot [color=black, dashed, forget plot]
  table[row sep=crcr]{
2	0.467575570413761\\
2	0.49051215927197\\
};
\addplot [color=black, dashed, forget plot]
  table[row sep=crcr]{
3	0.623462628487737\\
3	0.654284681684086\\
};
\addplot [color=black, dashed, forget plot]
  table[row sep=crcr]{
1	0.58567569606099\\
1	0.629011280676876\\
};
\addplot [color=black, dashed, forget plot]
  table[row sep=crcr]{
2	0.442633963453697\\
2	0.44463826480751\\
};
\addplot [color=black, dashed, forget plot]
  table[row sep=crcr]{
3	0.504469275444269\\
3	0.567346585049358\\
};
\addplot [color=black, forget plot]
  table[row sep=crcr]{
0.8875	0.673586347345813\\
1.1125	0.673586347345813\\
};
\addplot [color=black, forget plot]
  table[row sep=crcr]{
1.8875	0.49051215927197\\
2.1125	0.49051215927197\\
};
\addplot [color=black, forget plot]
  table[row sep=crcr]{
2.8875	0.654284681684086\\
3.1125	0.654284681684086\\
};
\addplot [color=black, forget plot]
  table[row sep=crcr]{
0.8875	0.58567569606099\\
1.1125	0.58567569606099\\
};
\addplot [color=black, forget plot]
  table[row sep=crcr]{
1.8875	0.442633963453697\\
2.1125	0.442633963453697\\
};
\addplot [color=black, forget plot]
  table[row sep=crcr]{
2.8875	0.504469275444269\\
3.1125	0.504469275444269\\
};
\addplot [color=blue, forget plot]
  table[row sep=crcr]{
0.775	0.629011280676876\\
0.775	0.658153960883433\\
1.225	0.658153960883433\\
1.225	0.629011280676876\\
0.775	0.629011280676876\\
};
\addplot [color=blue, forget plot]
  table[row sep=crcr]{
1.775	0.44463826480751\\
1.775	0.467575570413761\\
2.225	0.467575570413761\\
2.225	0.44463826480751\\
1.775	0.44463826480751\\
};
\addplot [color=blue, forget plot]
  table[row sep=crcr]{
2.775	0.567346585049358\\
2.775	0.623462628487737\\
3.225	0.623462628487737\\
3.225	0.567346585049358\\
2.775	0.567346585049358\\
};
\addplot [color=red, forget plot]
  table[row sep=crcr]{
0.775	0.651871679508622\\
1.225	0.651871679508622\\
};
\addplot [color=red, forget plot]
  table[row sep=crcr]{
1.775	0.456719463538129\\
2.225	0.456719463538129\\
};
\addplot [color=red, forget plot]
  table[row sep=crcr]{
2.775	0.613426986539909\\
3.225	0.613426986539909\\
};
\addplot [color=black, draw=none, mark=+, mark options={solid, red}, forget plot]
  table[row sep=crcr]{
1	0.50152480055018\\
1	0.502085679973765\\
1	0.502098429841572\\
1	0.502283161087677\\
1	0.502576879968564\\
1	0.584880986187685\\
};
\addplot [color=black, draw=none, mark=+, mark options={solid, red}, forget plot]
  table[row sep=crcr]{
nan	nan\\
};
\addplot [color=black, draw=none, mark=+, mark options={solid, red}, forget plot]
  table[row sep=crcr]{
nan	nan\\
};
\end{axis}
\end{tikzpicture}
		\caption{Scores for pre-processed and unprocessed input data.}
		\label{fig:boxplot-data-source}
	\end{subfigure}
	\hspace{1em}
	\begin{subfigure}[t]{0.45\linewidth}
\begin{tikzpicture}
\footnotesize
\begin{axis}[
width=0.8\linewidth,
height=0.6\linewidth,
scale only axis,
unbounded coords=jump,
xmin=0.5,
xmax=6.5,
xtick={1,2,3,4,5,6,7,8,9},
xticklabels={{3},{5},{7},{8},{9},{10},{15},{20},{30}},
xlabel style={font=\color{white!15!black}},
xlabel={length of Bakis structure},
ymin=0.492921723210398,
ymax=0.682189424685595,
ylabel style={font=\color{white!15!black}},
axis background/.style={fill=white},
xmajorgrids,
ymajorgrids
]
\addplot [color=black, dashed, forget plot]
  table[row sep=crcr]{
1	0.502356590807899\\
1	0.502576879968564\\
};
\addplot [color=black, dashed, forget plot]
  table[row sep=crcr]{
2	0.628954662421089\\
2	0.629124517188451\\
};
\addplot [color=black, dashed, forget plot]
  table[row sep=crcr]{
3	0.653695069437886\\
3	0.654527913676702\\
};
\addplot [color=black, dashed, forget plot]
  table[row sep=crcr]{
4	0.652348614447391\\
4	0.653676204894719\\
};
\addplot [color=black, dashed, forget plot]
  table[row sep=crcr]{
5	0.659614926419841\\
5	0.662838282777655\\
};
\addplot [color=black, dashed, forget plot]
  table[row sep=crcr]{
6	0.660305797333103\\
6	0.662419915099715\\
};
\addplot [color=black, dashed, forget plot]
  table[row sep=crcr]{
7	0.638248648589537\\
7	0.654138283697392\\
};
\addplot [color=black, dashed, forget plot]
  table[row sep=crcr]{
8	0.660698959052753\\
8	0.664995319445422\\
};
\addplot [color=black, dashed, forget plot]
  table[row sep=crcr]{
9	0.671365648874727\\
9	0.673586347345813\\
};
\addplot [color=black, dashed, forget plot]
  table[row sep=crcr]{
1	0.50152480055018\\
1	0.501945460117869\\
};
\addplot [color=black, dashed, forget plot]
  table[row sep=crcr]{
2	0.627478296712955\\
2	0.627968162908501\\
};
\addplot [color=black, dashed, forget plot]
  table[row sep=crcr]{
3	0.647242114800056\\
3	0.647665608288825\\
};
\addplot [color=black, dashed, forget plot]
  table[row sep=crcr]{
4	0.641117384871545\\
4	0.645353470556868\\
};
\addplot [color=black, dashed, forget plot]
  table[row sep=crcr]{
5	0.652933645743269\\
5	0.652933645743269\\
};
\addplot [color=black, dashed, forget plot]
  table[row sep=crcr]{
6	0.641005184461338\\
6	0.652341331997659\\
};
\addplot [color=black, dashed, forget plot]
  table[row sep=crcr]{
7	0.584880986187685\\
7	0.585477018592664\\
};
\addplot [color=black, dashed, forget plot]
  table[row sep=crcr]{
8	0.653673140820886\\
8	0.654501679842084\\
};
\addplot [color=black, dashed, forget plot]
  table[row sep=crcr]{
9	0.665095588717937\\
9	0.667158792827756\\
};
\addplot [color=black, forget plot]
  table[row sep=crcr]{
0.875	0.502576879968564\\
1.125	0.502576879968564\\
};
\addplot [color=black, forget plot]
  table[row sep=crcr]{
1.875	0.629124517188451\\
2.125	0.629124517188451\\
};
\addplot [color=black, forget plot]
  table[row sep=crcr]{
2.875	0.654527913676702\\
3.125	0.654527913676702\\
};
\addplot [color=black, forget plot]
  table[row sep=crcr]{
3.875	0.653676204894719\\
4.125	0.653676204894719\\
};
\addplot [color=black, forget plot]
  table[row sep=crcr]{
4.875	0.662838282777655\\
5.125	0.662838282777655\\
};
\addplot [color=black, forget plot]
  table[row sep=crcr]{
5.875	0.662419915099715\\
6.125	0.662419915099715\\
};
\addplot [color=black, forget plot]
  table[row sep=crcr]{
6.875	0.654138283697392\\
7.125	0.654138283697392\\
};
\addplot [color=black, forget plot]
  table[row sep=crcr]{
7.875	0.664995319445422\\
8.125	0.664995319445422\\
};
\addplot [color=black, forget plot]
  table[row sep=crcr]{
8.875	0.673586347345813\\
9.125	0.673586347345813\\
};
\addplot [color=black, forget plot]
  table[row sep=crcr]{
0.875	0.50152480055018\\
1.125	0.50152480055018\\
};
\addplot [color=black, forget plot]
  table[row sep=crcr]{
1.875	0.627478296712955\\
2.125	0.627478296712955\\
};
\addplot [color=black, forget plot]
  table[row sep=crcr]{
2.875	0.647242114800056\\
3.125	0.647242114800056\\
};
\addplot [color=black, forget plot]
  table[row sep=crcr]{
3.875	0.641117384871545\\
4.125	0.641117384871545\\
};
\addplot [color=black, forget plot]
  table[row sep=crcr]{
4.875	0.652933645743269\\
5.125	0.652933645743269\\
};
\addplot [color=black, forget plot]
  table[row sep=crcr]{
5.875	0.641005184461338\\
6.125	0.641005184461338\\
};
\addplot [color=black, forget plot]
  table[row sep=crcr]{
6.875	0.584880986187685\\
7.125	0.584880986187685\\
};
\addplot [color=black, forget plot]
  table[row sep=crcr]{
7.875	0.653673140820886\\
8.125	0.653673140820886\\
};
\addplot [color=black, forget plot]
  table[row sep=crcr]{
8.875	0.665095588717937\\
9.125	0.665095588717937\\
};
\addplot [color=blue, forget plot]
  table[row sep=crcr]{
0.75	0.501945460117869\\
0.75	0.502356590807899\\
1.25	0.502356590807899\\
1.25	0.501945460117869\\
0.75	0.501945460117869\\
};
\addplot [color=blue, forget plot]
  table[row sep=crcr]{
1.75	0.627968162908501\\
1.75	0.628954662421089\\
2.25	0.628954662421089\\
2.25	0.627968162908501\\
1.75	0.627968162908501\\
};
\addplot [color=blue, forget plot]
  table[row sep=crcr]{
2.75	0.647665608288825\\
2.75	0.653695069437886\\
3.25	0.653695069437886\\
3.25	0.647665608288825\\
2.75	0.647665608288825\\
};
\addplot [color=blue, forget plot]
  table[row sep=crcr]{
3.75	0.645353470556868\\
3.75	0.652348614447391\\
4.25	0.652348614447391\\
4.25	0.645353470556868\\
3.75	0.645353470556868\\
};
\addplot [color=blue, forget plot]
  table[row sep=crcr]{
4.75	0.652933645743269\\
4.75	0.659614926419841\\
5.25	0.659614926419841\\
5.25	0.652933645743269\\
4.75	0.652933645743269\\
};
\addplot [color=blue, forget plot]
  table[row sep=crcr]{
5.75	0.652341331997659\\
5.75	0.660305797333103\\
6.25	0.660305797333103\\
6.25	0.652341331997659\\
5.75	0.652341331997659\\
};
\addplot [color=blue, forget plot]
  table[row sep=crcr]{
6.75	0.585477018592664\\
6.75	0.638248648589537\\
7.25	0.638248648589537\\
7.25	0.585477018592664\\
6.75	0.585477018592664\\
};
\addplot [color=blue, forget plot]
  table[row sep=crcr]{
7.75	0.654501679842084\\
7.75	0.660698959052753\\
8.25	0.660698959052753\\
8.25	0.654501679842084\\
7.75	0.654501679842084\\
};
\addplot [color=blue, forget plot]
  table[row sep=crcr]{
8.75	0.667158792827756\\
8.75	0.671365648874727\\
9.25	0.671365648874727\\
9.25	0.667158792827756\\
8.75	0.667158792827756\\
};
\addplot [color=red, forget plot]
  table[row sep=crcr]{
0.75	0.502098429841572\\
1.25	0.502098429841572\\
};
\addplot [color=red, forget plot]
  table[row sep=crcr]{
1.75	0.628137997358674\\
2.25	0.628137997358674\\
};
\addplot [color=red, forget plot]
  table[row sep=crcr]{
2.75	0.648374357309548\\
3.25	0.648374357309548\\
};
\addplot [color=red, forget plot]
  table[row sep=crcr]{
3.75	0.651837274718963\\
4.25	0.651837274718963\\
};
\addplot [color=red, forget plot]
  table[row sep=crcr]{
4.75	0.658199799829008\\
5.25	0.658199799829008\\
};
\addplot [color=red, forget plot]
  table[row sep=crcr]{
5.75	0.658108121937859\\
6.25	0.658108121937859\\
};
\addplot [color=red, forget plot]
  table[row sep=crcr]{
6.75	0.632716164775247\\
7.25	0.632716164775247\\
};
\addplot [color=red, forget plot]
  table[row sep=crcr]{
7.75	0.655866408761683\\
8.25	0.655866408761683\\
};
\addplot [color=red, forget plot]
  table[row sep=crcr]{
8.75	0.669881680792694\\
9.25	0.669881680792694\\
};
\addplot [color=black, draw=none, mark=+, mark options={solid, red}, forget plot]
  table[row sep=crcr]{
nan	nan\\
};
\addplot [color=black, draw=none, mark=+, mark options={solid, red}, forget plot]
  table[row sep=crcr]{
nan	nan\\
};
\addplot [color=black, draw=none, mark=+, mark options={solid, red}, forget plot]
  table[row sep=crcr]{
nan	nan\\
};
\addplot [color=black, draw=none, mark=+, mark options={solid, red}, forget plot]
  table[row sep=crcr]{
nan	nan\\
};
\addplot [color=black, draw=none, mark=+, mark options={solid, red}, forget plot]
  table[row sep=crcr]{
5	0.641194843485208\\
};
\addplot [color=black, draw=none, mark=+, mark options={solid, red}, forget plot]
  table[row sep=crcr]{
nan	nan\\
};
\addplot [color=black, draw=none, mark=+, mark options={solid, red}, forget plot]
  table[row sep=crcr]{
nan	nan\\
};
\addplot [color=black, draw=none, mark=+, mark options={solid, red}, forget plot]
  table[row sep=crcr]{
nan	nan\\
};
\addplot [color=black, draw=none, mark=+, mark options={solid, red}, forget plot]
  table[row sep=crcr]{
nan	nan\\
};
\end{axis}
\end{tikzpicture}
		\caption{
			Scores for different lengths $I$ of the Bakis sub-models.
		}
		\label{fig:boxplot-bakis}
	\end{subfigure}
	\caption[$F_1$ scores evaluating input data and HMM model.]{$F_1$ scores evaluating input data and HMM model.}
\end{figure}
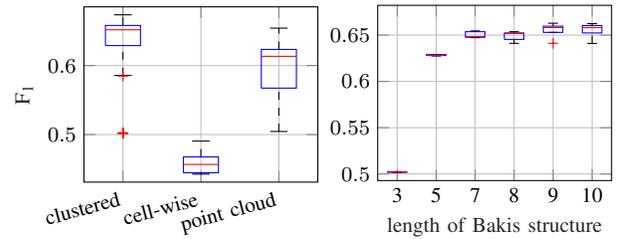
Figure \ref{fig:boxplot-data-source} depicts results for different pre- and unprocessed SOGMs.
Compared to a training on cell-wise data, the results based on the supercell segmentation yield significantly higher $F_1$ scores.
This shows the effectiveness of the supercell segmentation and confirms the motivation of using it as a de-noising method.
The lower scores for the point cloud representation are most likely explained by the inaccuracies introduced by the reduction of the SOGM to topometric maps, which discard the shape of supercells (cf. \cite{Korthals2017a}).
For further evaluations only input data from the the supercell segmented SOGMs is considered.

A baseline classifier which chooses the most frequent class yields an averaged $F_1$ of $0.29$, a random classifier based on the distribution of class labels in the training data offers a slightly higher $F_1$ score of $0.33$.
Classification using a k-means clustering approach achieves a $F_1$ score of $0.58$.
The highest $F_1$ score of $0.66$ is achieved by the trained HMM model initialized with the manually chosen means, a Bakis length greater than 8 and initial variances of 0.1 (cf. \autoref{fig:boxplot-bakis}).
Generally, the trained HMM models perform equally well or better than the k-means clustering with the exception of the model with three states per Bakis sub-model.
\section{CONCLUSION \& FURTHER PROSPECTS }
The presented work shows the advantage of applying a clustering approach to refine SOGMs.
The application of HMMs, which respects a sequential path nature, to infer the hidden true state from the SOGM outperforms common classifier approaches.
These results give a good baseline to extend the approach of a SOGM as environmental representation to perform  informed path planning, POMDP, or (Deep)-Q-Learning on multi-modal data.
\section*{ACKNOWLEDGMENT}
\small
This research was supported by the Cluster of Excellence Cognitive Interaction Technology 'CITEC' (EXC 277) at Bielefeld University and by the Federal Ministry of Education and Research under grant number 57388272. The responsibility for the content of this publication lies with the author.
\normalsize
\bibliographystyle{IEEEtran}
\bibliography{root}

\end{document}

%% file: graph_object.tex
%
%
\begin{tikzpicture}
\footnotesize
\begin{axis}[%
width=0.4\linewidth,
height=0.3\linewidth,
scale only axis,
xmin=0,
xmax=0.55,
xlabel style={font=\color{white!15!black}},
xlabel={Distance in meter},
ymin=0,
ymax=1,
ylabel style={font=\color{white!15!black}},
ylabel={$\text{P(m}\text{)}$},
axis background/.style={fill=white},
xmajorgrids,
ymajorgrids,
legend style={legend cell align=left, align=left, draw=white!15!black},
legend pos=south east
]
\addplot [color=blue, mark=x, mark options={solid, blue}]
  table[row sep=crcr]{%
0	0.00035697064595297\\
0.00999999046325684	0.00035697064595297\\
0.0223606582731009	0.00035697064595297\\
0.0316227450966835	0.00035697064595297\\
0.0412310175597668	0.00035697064595297\\
0.0538515970110893	0.00035697064595297\\
0.063245490193367	0.00035697064595297\\
0.0728010311722755	0.00035697064595297\\
0.0824620351195335	0.00035697064595297\\
0.094868466258049	0.00035697064595297\\
0.10440319031477	0.00035697064595297\\
0.1140176653862	0.00035697064595297\\
0.126491218805313	0.0159063916653395\\
0.136014804244041	0.0229773707687855\\
0.145602285861969	0.0474258735775948\\
0.158113956451416	0.256832003593445\\
0.167630612850189	0.256832003593445\\
0.177200511097908	0.256832003593445\\
0.189736738801003	0.468790620565414\\
0.199248656630516	0.577495336532593\\
0.208806186914444	0.577495336532593\\
0.221359491348267	0.679178714752197\\
0.230867967009544	0.76629364490509\\
0.240416333079338	0.76629364490509\\
0.250000029802322	0.835483551025391\\
0.262488096952438	0.88079708814621\\
0.272029399871826	0.88079708814621\\
0.281602561473846	0.874077260494232\\
0.294108927249908	0.874077260494232\\
0.303644627332687	0.874077260494232\\
0.313209265470505	0.867035746574402\\
0.32573002576828	0.859663724899292\\
0.335261136293411	0.859663724899292\\
0.344818830490112	0.859663724899292\\
0.357351392507553	0.851952791213989\\
0.366878747940063	0.843895077705383\\
0.376430630683899	0.843895077705383\\
0.388973027467728	0.835483551025391\\
0.39849716424942	0.835483551025391\\
0.408044099807739	0.817574501037598\\
0.417612224817276	0.817574501037598\\
0.430116355419159	0.808067202568054\\
0.439659029245377	0.798186779022217\\
0.449221611022949	0.798186779022217\\
0.461735904216766	0.798186779022217\\
0.47127491235733	0.798186779022217\\
0.480832636356354	0.787931203842163\\
0.493355870246887	0.76629364490509\\
0.502891659736633	0.76629364490509\\
};
\addlegendentry{Anomaly}

\addplot [color=green, mark=o, mark options={solid, green}]
  table[row sep=crcr]{%
0	0.0566524267196655\\
0.00999999046325684	0.0566524267196655\\
0.0223606582731009	0.0566524267196655\\
0.0316227450966835	0.0566524267196655\\
0.0412310175597668	0.0566524267196655\\
0.0538515970110893	0.0566524267196655\\
0.063245490193367	0.0566524267196655\\
0.0728010311722755	0.0566524267196655\\
0.0824620351195335	0.0566524267196655\\
0.094868466258049	0.0566524267196655\\
0.10440319031477	0.0566524267196655\\
0.1140176653862	0.00035697064595297\\
0.126491218805313	0.00035697064595297\\
0.136014804244041	0.00035697064595297\\
0.145602285861969	0.00035697064595297\\
0.158113956451416	0.5\\
0.167630612850189	0.348645120859146\\
0.177200511097908	0.348645120859146\\
0.189736738801003	0.5\\
0.199248656630516	0.5\\
0.208806186914444	0.5\\
0.221359491348267	0.5\\
0.230867967009544	0.5\\
0.240416333079338	0.5\\
0.250000029802322	0.5\\
0.262488096952438	0.5\\
0.272029399871826	0.5\\
0.281602561473846	0.5\\
0.294108927249908	0.5\\
0.303644627332687	0.5\\
0.313209265470505	0.5\\
0.32573002576828	0.5\\
0.335261136293411	0.5\\
0.344818830490112	0.5\\
0.357351392507553	0.5\\
0.366878747940063	0.5\\
0.376430630683899	0.5\\
0.388973027467728	0.5\\
0.39849716424942	0.5\\
0.408044099807739	0.5\\
0.417612224817276	0.5\\
0.430116355419159	0.5\\
0.439659029245377	0.5\\
0.449221611022949	0.5\\
0.461735904216766	0.5\\
0.47127491235733	0.5\\
0.480832636356354	0.5\\
0.493355870246887	0.5\\
0.502891659736633	0.5\\
};
\addlegendentry{Corner}

\addplot [color=red, mark=*, mark options={solid, red}, mark size=1pt]
  table[row sep=crcr]{%
0	0.119202919304371\\
0.00999999046325684	0.119202919304371\\
0.0223606582731009	0.119202919304371\\
0.0316227450966835	0.119202919304371\\
0.0412310175597668	0.119202919304371\\
0.0538515970110893	0.119202919304371\\
0.063245490193367	0.119202919304371\\
0.0728010311722755	0.119202919304371\\
0.0824620351195335	0.119202919304371\\
0.094868466258049	0.119202919304371\\
0.10440319031477	0.119202919304371\\
0.1140176653862	0.99193799495697\\
0.126491218805313	0.99193799495697\\
0.136014804244041	0.99193799495697\\
0.145602285861969	0.99193799495697\\
0.158113956451416	0.5\\
0.167630612850189	0.348645120859146\\
0.177200511097908	0.00035697064595297\\
0.189736738801003	0.00035697064595297\\
0.199248656630516	0.5\\
0.208806186914444	0.5\\
0.221359491348267	0.5\\
0.230867967009544	0.5\\
0.240416333079338	0.5\\
0.250000029802322	0.5\\
0.262488096952438	0.5\\
0.272029399871826	0.5\\
0.281602561473846	0.5\\
0.294108927249908	0.5\\
0.303644627332687	0.5\\
0.313209265470505	0.5\\
0.32573002576828	0.5\\
0.335261136293411	0.5\\
0.344818830490112	0.5\\
0.357351392507553	0.5\\
0.366878747940063	0.5\\
0.376430630683899	0.5\\
0.388973027467728	0.5\\
0.39849716424942	0.5\\
0.408044099807739	0.5\\
0.417612224817276	0.5\\
0.430116355419159	0.5\\
0.439659029245377	0.5\\
0.449221611022949	0.5\\
0.461735904216766	0.5\\
0.47127491235733	0.5\\
0.480832636356354	0.5\\
0.493355870246887	0.5\\
0.502891659736633	0.5\\
};
\addlegendentry{Obstacle}

\end{axis}
\end{tikzpicture}%